\documentclass[journal]{IEEEtran}

\pdfoutput=1 

\usepackage[utf8]{inputenc}
\usepackage[T1]{fontenc}
\usepackage{textcomp}
\usepackage{gensymb}
\usepackage{amsmath}
\usepackage{makecell}
\usepackage[utf8]{inputenc}
\usepackage{booktabs}
\usepackage{longtable}
\usepackage{tabularx}
\usepackage{amssymb}
\usepackage{threeparttable}
\usepackage{array,multirow,graphicx}

\ifCLASSINFOpdf
\else
\fi

\begin{document}
\newpage 
\onecolumn %
\pagestyle{empty} %
  {\large  %
This work has been submitted to the IEEE for possible publication. 
Copyright may be transferred without notice, after which this version may no 
longer be accessible.}
\newpage
\twocolumn %
\setcounter{page}{1} %
\pagestyle{headings}

\title{PanorAMS: Automatic Annotation for \\ Detecting Objects in Urban Context}

\author{Inske Groenen,
        Stevan Rudinac,
        and~Marcel Worring,~\IEEEmembership{Senior Member,~IEEE}}

\maketitle

\begin{abstract}
 Large collections of geo-referenced panoramic images are freely available for cities across the globe, as well as detailed maps with location and meta-data on a great variety of urban objects. They provide a potentially rich source of information on urban objects, but manual annotation for object detection is costly, laborious and difficult. Can we utilize such multimedia sources to automatically annotate street level images  as an inexpensive alternative to manual labeling? With the PanorAMS framework we introduce a method to automatically generate bounding box annotations for panoramic images based on urban context information. Following this method, we acquire large-scale, albeit noisy, annotations for an urban dataset solely from open data sources in a fast and automatic manner. The dataset covers the City of Amsterdam and includes over 14 million noisy bounding box annotations of 22 object categories present in 771,299 panoramic images. For many objects further fine-grained information is available, obtained from geospatial meta-data, such as \emph{building value}, \emph{function} and \emph{average surface area}. Such information would have been difficult, if not impossible, to acquire via manual labeling based on the image alone. For detailed evaluation, we introduce an efficient crowdsourcing protocol for bounding box annotations in panoramic images, which we deploy to acquire 147,075 ground-truth object annotations for a subset of 7,348 images, the PanorAMS-clean dataset. For our PanorAMS-noisy dataset, we provide an extensive analysis of the noise and how different types of noise affect image classification and object detection performance. We make both datasets, PanorAMS-noisy and PanorAMS-clean, benchmarks and tools presented in this paper openly available.
\end{abstract}


\IEEEpeerreviewmaketitle

\section{Introduction}

Detecting urban objects in geo-referenced images is relevant to many applications, ranging from autonomous driving \cite{kitti, 10.1145/3126686.3126727, 10.1145/2964284.2967294} to urban planning and analysing city life \cite{urban_object_detection_kit, urban_footpath_dataset}. While a large amount of geo-referenced image data is available for cities around the world, the tremendous cost and labour involved in acquiring suitable annotations is a major limiting factor in training current generic state-of-the-art object detectors for the specifics of urban datasets. Most existing urban object detection datasets, such as KITTI \cite{kitti}, required human annotators to manually draw precise bounding boxes in thousands of images. To bring forward the various applications, more efficient ways of annotation are needed.

\begin{figure}
  \centering
  \includegraphics[width=0.48\textwidth]{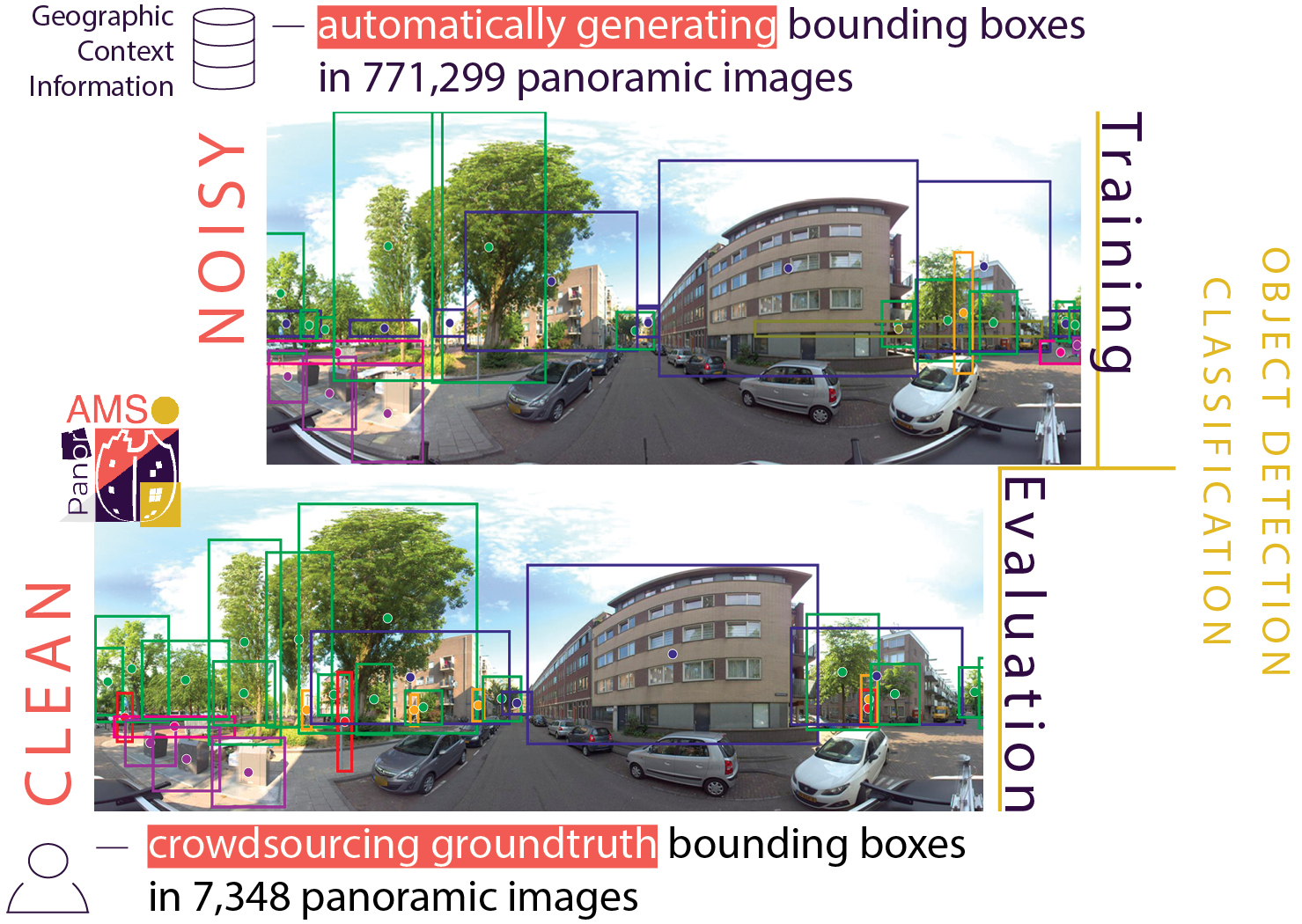}
  \caption{Overview of our PanorAMS framework. We use open-source geographic context information to automatically generate bounding boxes in 360{\textdegree} panoramic images of the City of Amsterdam. By doing so, we acquire the large-scale PanorAMS-noisy dataset. We use the TrainVal subset of this dataset to train image classification and object detection networks. We implement an efficient crowd-sourcing protocol specific for 360{\textdegree} images to obtain accurate bounding box annotation for a subset of the images. This way we acquire the PanorAMS-clean dataset, which can be used as a standalone detection dataset with clean annotations or as a means to evaluate the performance of networks trained in our noisy setting.}
  \label{fig:panorams_brief_overview}
\end{figure}

Recent years have seen a noticeable increase in research efforts to elevate the annotation burden in object detection. Many such efforts have focused on weak supervision settings, such as using only image class labels \cite{7581027, 10.1145/3394171.3413835, 10.1145/3123266.3123455}, scribble \cite{Zhang_2020_CVPR} and click supervision \cite{Papadopoulos_2017_CVPR}. These methods still require some form of accurate labeling.
In an attempt to completely eliminate the need for manual annotation, detectors have been trained on queried images from the web \cite{8489885, 7882707, DBLP:journals/tkde/YaoZSLZZS20}. This setting is even more challenging as not all retrieved images will be relevant to the query, thus, introducing class label noise into the dataset. These methods are further limited to one central object per image, while the urban context often involves analysing crowded street imagery. On the other hand, a key characteristic of the urban context is that modern cities nowadays are largely mapped out. Maps with detailed object information are a potential rich source to automatically annotate geo-referenced images. 

Urban multimedia sources that contain relevant information for automatic image annotation include geospatial object information, elevation map data and human knowledge of the city. By combining these multimedia sources, essential 3D real-world object measurements can be extracted. The width and distance of an object in relation to a geo-referenced image can, for instance, be determined on the basis of object coordinates contained in Geographic Information Systems (GIS). The main challenge in converting the obtained 3D measurements to 2D street level image coordinates is the drastic change in viewpoint. While the conversion itself can be done using a simple camera model \cite{gis_eccv14,geosemantic_segmentation,urbantrees}, it is difficult to determine which objects are occluded from a street level viewpoint, as opposed to aerial where occlusions are minimal but little information on 3D structure is present. Another problem is dealing with measurement inaccuracies which arise from uncertainties in the registered image location, geographic object coordinates, and real-world (estimated) measurements of objects, and cause the image location of generated annotations to be imprecise, thus, introducing bounding box location noise. Previous work required neural networks to handle occlusion and location noise, and has been limited either in the diversity of objects \cite{geosemantic_segmentation, urbantrees,wang2017torontocity,mapsupervised_road_detection, mattyus2016hdmaps}, the number of images \cite{gis_eccv14,geosemantic_segmentation,Wang_2015_CVPR}, or the variety of data sources \cite{gis_eccv14, geosemantic_segmentation, mapsupervised_road_detection} involved. In this work, we use multiple open-source multimedia sources to generate millions of bounding boxes in hundreds of thousands images for a variety of object classes and deal with occlusion through simple geometric reasoning, i.e. based on bounding box overlap and real-world distance between objects and the image location, without the use of dedicated networks.

Following our automatic annotation pipeline, we create a large-scale dataset with bounding box annotations solely from open data sources in a fast and inexpensive manner. The PanorAMS-noisy dataset covers the entire land area of 219.5 km$^2$ of the City of Amsterdam and contains 14,821,852 labeled object instances in 771,299 panoramic street level images. It contains 22 object categories commonly found in street scenery (e.g. \emph{building}, \emph{tree}, and \emph{lamp post}) with further fine-grained information obtained from geospatial meta-data for certain classes (e.g. \emph{building value}, \emph{function} and \emph{average surface area}). Thus, by capitalising on the wealth of contextual information available within the urban setting we not only move away from manual labeling, we also obtain richer object information compared to previous urban detection datasets. 

To evaluate the quality of our automatically generated bounding boxes, we crowdsource ground-truth annotations for a subset of the images contained in PanorAMS-noisy. For this, we implement an efficient crowdsourcing protocol using the generated boxes as a starting point. In the interest of minimizing the required annotation time, our protocol focuses on preventing task-switching, which is cognitively demanding, and keeping the necessary mouse and eye movements to a minimum. We divide our protocol into three sub-tasks based on \cite{crowdsourcing_annotations_object_detection} and implement extreme clicking, a fast box annotation technique by clicking the object's four extreme points (i.e. top, bottom, left-most, right-most) \cite{extreme_clicking}. It further includes a specific method to label an object that is present on both the left- and right-side of the image (due to the nature of 360\textdegree{} panoramic images), in which case two boxes need to be be labeled as belonging to the same object. Following our protocol, we crowdsource 147,075 ground-truth bounding box annotations for a subset of 7,348 images of the PanorAMS-noisy dataset, resulting in the PanorAMS-clean dataset.

Finally, we provide extensive analysis of classification and detection performance in our noisy label setting. By analyzing how different types of noise affect network performance, we provide valuable insights into training Deep Neural Networks (DNNs) on a real-world dataset with annotations involving both class and bounding box location noise.

To summarize, in this paper we introduce the PanorAMS framework, which encompasses the following contributions:

\begin{itemize}
    \item a method to automatically generate bounding box annotations for a variety of object classes in panoramic street-level images based on urban context information,
    \item a protocol to efficiently crowdsource ground-truth bounding box annotations in 360\textdegree{} images with noisy boxes as a starting point,
    \item a large-scale urban dataset containing 14,821,852 noisy bounding box annotations in 771,299 images, associated with rich object meta-data (PanorAMS-noisy), and 147,075 crowdsourced ground-truth annotations for a subset of 7,348 images (PanorAMS-clean),
    \item detailed analysis of the effect of different types of label noise on classification and detection performance.
\end{itemize}

\begin{table*}[tb!]
\caption{Comparison of the PanorAMS datasets with other object detection datasets concerning street imagery}
\centering
\label{tab:datasets_comparison}
\begin{threeparttable}
\begin{tabular}{lrrrlcccc}
\toprule
 & {\# Images} & {\# Classes} & {\# Instances} & \thead{Horizontal \\ Field of View} & \thead{Image \\ Distortion} & \thead{Map \\ Data} & \thead{Meta \\ Data} & \thead{Automatic \\ Annotation} \\
\midrule
KITTI \cite{kitti} & 14,999 & 3 &  80,256 & \quad< 180\degree & {$\times$} &  {$\times$}  & \quad{$\times$} & {$\times$} \\
SUN2012\tnote{*} \space \cite{sun} & 16,873 & 3,819\tnote{**} & 249,522 & \quad< 180\degree & {$\times$} & {$\times$} & \quad{$\times$} &  {$\times$}  \\
Cityscapes\tnote{*} \space\cite{cityscapes} & 25,000 & 30 & 65,385 & \quad< 180\degree & {$\times$} &  {$\times$} & \quad{$\times$}  & {$\times$} \\
Mapillary Vistas\tnote{*} \space \cite{Mapillary_2017_ICCV}  & 25,000 & 37 & {$\backsim$} 400,000 & \quad< 180\degree & {$\times$} & {$\times$} &  \quad{$\times$}  & {$\times$}  \\
ADE20K\tnote{*} \space \cite{ADE20K_IJCV} & 22,210 & 2,693\tnote{**} & 434,826 & \quad< 180\degree & {$\times$} & {$\times$} &  \quad{$\times$}  & {$\times$}  \\
WoodScape \cite{woodscape_2019_ICCV} & 10,000 & 7 & 67,706 & \quad\quad360\degree & Fisheye & {$\times$} &  \quad{$\times$} & {$\times$} \\
Waymo Open \cite{waymo_CVPR_2020} & {$\backsim$} 200,000 & 4 & {$\backsim$} 9,900,000 & \quad{$\backsim$} 225\degree & {$\times$} & {$\times$} &  \quad {$\times$} & {$\times$} \\
PanorAMS-clean (ours) & 7,348 & 22 & 147,075 & \quad\quad360\degree  & Panorama & {\checkmark} &  \quad{\checkmark}\tnote{***} & {$\times$} \\
PanorAMS-noisy (ours) & 771,299 & 22  & 14,821,852 & \quad\quad360\degree & Panorama & {\checkmark} &  \quad {\checkmark} & {\checkmark} \\
\bottomrule
\end{tabular}
\begin{tablenotes}
\item[*] bounding boxes can be obtained from instance-level semantic segmentations
\item[**] dataset includes both street and indoor imagery, which explains the large number of classes compared to others
\item[***] metadata is linked to bounding boxes based on the PanorAMS-noisy dataset through automatic annotation 
\end{tablenotes}
\end{threeparttable}
\end{table*}

\section{Related work}

In this section, we first address 
previous work on using urban context to label images. We then provide a comprehensive overview of relevant existing datasets and their limitations, followed by relevant studies on learning from noisy labels. Finally, we address related work on acquiring ground-truth annotations via crowdsourcing.

\subsection{Automatic annotation based on urban context}

Geographic context information has previously been used to acquire image labels in an automatic way. One such approach is to label geo-referenced images without image tags based on image similarity with a set of user tagged images geo-referenced in the same area \cite{geobased_automatic_annotation}. Previous work on object detection has mostly used geospatial information as a way to obtain object priors \cite{gis_eccv14, Wang_2015_CVPR}, rather than using it as a way to obtain bounding boxes on which object detectors can be trained directly. One exception is TorontoCity \cite{wang2017torontocity}, which uses geospatial data of buildings and streets to obtain ground-truth object segmentations. TorontoCity does require specific algorithms to align geospatial data and image data due to a large error margin on the geolocation of images. While \cite{urbantrees} use a public tree inventory as the basis to create their dataset containing bounding box annotations of different tree species, a subset of crowdsourced annotations are still required to automatically label the remaining images.
Notably, Ardeshir et al. \cite{gis_eccv14} demonstrate how to automatically acquire bounding box annotations in street level images based on object distance measurements from geospatial information and object height and width measurements that were estimated on the basis of city observations. This is done only for a limited amount of images and object classes. A graph matching process with additional computational cost is required to deal with the noise associated with these automatically generated boxes in order to perform object detection.

\subsection{Urban classification and object detection datasets}

Popular urban benchmark datasets for classification include SUN \cite{sun} and Places365 \cite{zhou2017places}. A popular benchmark dataset for scene understanding in panoramic images is the SUN360 dataset \cite{sun360}. Most existing urban detection datasets, including KITTI \cite{kitti}, Cityscapes \cite{cityscapes}, Mapillary Vistas \cite{Mapillary_2017_ICCV}, and Waymo Open \cite{waymo_CVPR_2020}, were created with autonomous driving in mind. As a result, the main classes in these datasets concern moving objects, such as vehicles and people, and the variety of classes with regard to static objects encountered within urban scenes is limited for most. OpenStreetMap (OSM), which is one of our sources to obtain geospatial object information, was consulted for the Mapillary Vistas dataset to determine which object categories to include. However, they do not use geospatial information in the actual labeling process \cite{Mapillary_2017_ICCV}. TorontoCity is a large-scale urban detection dataset where geospatial information was used as a way to generate labels in an automatic way \cite{wang2017torontocity}. However, the authors do so for only two object categories, namely buildings and streets, and due to license restrictions the dataset is not publicly available. In this paper, we demonstrate how geospatial information from open data sources may be used to label 22 object classes selected and frequently used by domain experts. To the best of our knowledge, PanorAMS-clean is the largest publicly available panorama dataset with accurate bounding box annotations. The WoodScape \cite{woodscape_2019_ICCV} dataset concerns images of a similarly large horizontal field of view, but a different type of distortion. Table~\ref{tab:datasets_comparison} provides a comparison of datasets concerning street imagery.

\subsection{Learning from noisy labels}

Studies on the dynamics of modern deep neural networks (DNNs) have indicated that DNNs learn simple patterns first, before gradually memorizing all samples \cite{dnn_memorization_pmlr2017}. This memorization effect, in combination with the large number of trainable parameters, cause DNNs to easily overfit to noisy labels, leading to poor generalization performance \cite{dnn_memorization_pmlr2017, zhang2016understanding}. One well-known technique to prevent DNNs from overfitting to noisy labels is the use of regularizers, such as weight decay, dropout \cite{dropout}, mixup \cite{mixup}, and gradient centralization \cite{gradient_cent}. We implement these techniques during our training. Another way to prevent memorization of noisy labels is to use noise robust loss functions. Softmax cross-entropy, commonly used in multi-class classification, was shown to be sensitive to label noise \cite{cce_sensitive2noise}. In multi-label classification, asymmetric loss (ASL) was recently introduced as a robust loss that discards possibly mislabeled examples, while addressing the high imbalance in negative and positive examples \cite{ASL}. We implement ASL as classification loss in our models.

When investigating the effect of image label noise, researchers often experiment on originally clean datasets with synthetically introduced noise, such as CIFAR-10, CIFAR-100 and Tiny ImageNet. The number of real-world noisily labeled datasets is limited. Three popular real-world benchmark datasets are Clothing1M \cite{clothing1m}, containing 14 fashion classes, Food-101N \cite{food-101N}, containing 101 food categories, and WebVision \cite{webvision}, covering 1,000 semantic concepts in web-crawled images from Flickr and Google. With PanorAMS-noisy we introduce a new large-scale, publicly available, real-world dataset that, apart from class label noise, also involves bounding box location noise.

\subsection{Crowdsourcing}
There are a couple of open-source applications available to crowdsource bounding box annotations, such as the VGG Image Annotator \cite{10.1145/3343031.3350535} and LabelMe \cite{labelme}. However, none of the available applications had all the functionalities we required. They do not allow for loading in our automatically generated bounding boxes nor do they have a particularly suited layout for panoramic images. We therefore chose to develop our own annotation tool based on best practices. Easy creation and adjustability of bounding boxes are key elements of a good annotation tool \cite{extreme_clicking, video_annotation_tool_lessons_learned}. We design our tool in such a way that required eye and mouse movements are kept to a minimum. We follow \cite{crowdsourcing_annotations_object_detection} as an authoritative source on crowdsourcing bounding box annotations, and divide our annotation protocol up into sub-tasks in a similar manner. Extreme clicking has proven to be a fast and accurate way of creating new bounding boxes \cite{extreme_clicking}. We implement this method in our tool. We further introduce the concept of linked bounding boxes that is specific to objects split across the left and right side of 360{\textdegree} images, whereby two bounding boxes are labeled as belonging to the same object.

\begin{figure*}
  \centering
  \includegraphics[width=\textwidth]{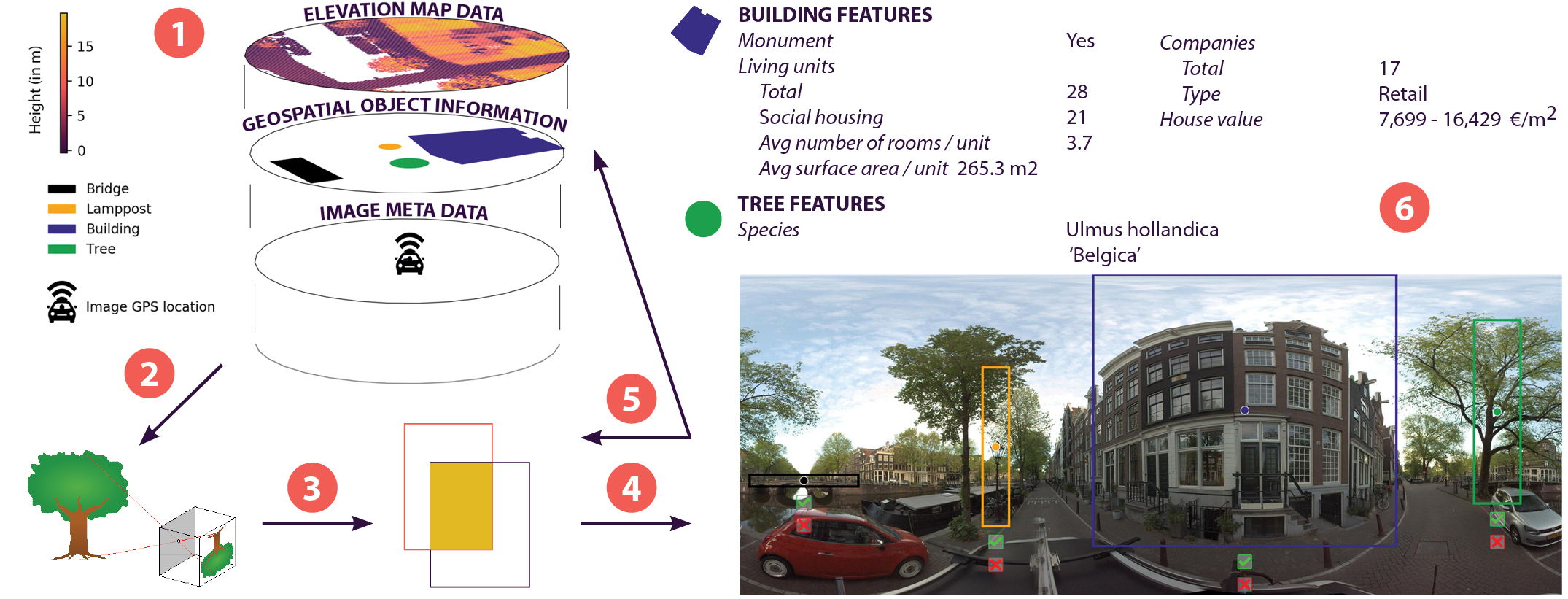}
  \caption{Overview of our pipeline to automatically generate noisy bounding box annotations in 360\textdegree{} street level images based on geographic context information. 
  (1) Based on city observations, geospatial object information, and elevation map data, acquire object attributes and 3D real-world measurements of all objects falling within a 150 meter radius of the image GPS location. (2) Convert this information to 2D image coordinates using the pinhole camera model in order to generate an initial set of bounding boxes. (3) Refine and filter the initial set of bounding boxes acquired during step 2 via geometric reasoning based on the percentage of overlap between boxes, the classes associated with overlapping boxes, and the real-world distance between overlapping objects and the camera. Urban knowledge is incorporated at this stage by optimizing these thresholds per class. (4) Map the final set of bounding boxes onto the image in order to qualitatively analyze the generated bounding boxes per class. (5) Optimize class rules, thresholds, and estimates of objects' real-world measurements by qualitative analysis of images and corresponding bounding boxes. (6) Link object metadata that is available from geospatial object information (e.g. building value) to the automatically generated bounding boxes. 
  }
    \label{fig:automatic_labeling}
\end{figure*}

\section{Using Geographic Context for Automatic Object Annotation: PanorAMS-noisy Dataset} \label{sec:panorams:noisy}
The high population density and great variety of object appearances across administrative areas make Amsterdam a uniquely challenging place of study (see Figure~\ref{fig:detection_qualitative} for some image examples of different geographic locations in Amsterdam). Amsterdam is the largest city of the Netherlands with a population of roughly 850,000 people. The PanorAMS-noisy dataset covers the entire land area of 219.5 km$^2$ falling within the official city boundaries of Amsterdam. It is a densely populated area with 5,296 inhabitants/km$^2$ and 2,715 houses/km$^2$ \cite{kerncijfers_AMS}. The municipality can be subdivided into different administrative areas, namely 99 quarters or 481 neighbourhoods \cite{Data_ams}. We acquire geographic context information from three multimedia sources: high-resolution panoramic images, elevation map data, and geospatial object information, which we will elaborate upon in the next section. Thereafter, we will describe how geographic context information can be used to generate automatic bounding box annotations.

\subsection{Multimedia sources}

\subsubsection{Panoramas} We collected 1,994,578 street level panoramic images of Amsterdam by utilizing the data portal of the City of Amsterdam \cite{Data_ams}. Moving vehicles with six cameras mounted on top captured these images over both land and water during the period 17-03-2016 to 23-01-2020 by simultaneously taking six photos and stitching these photos together to get the final 360\textdegree{} equirectangular image projection. For certain geographic locations the initial set contains images spanning multiple years, while for other locations this is not the case. We need to filter the images to counter oversampling of certain geographic locations, while at the same time ensuring that our set contains the most recent images available for each location. As the moving vehicle captured images at a distance interval of 5 meter from one image to the next, we established that the images in our dataset have to be at least 2.5 meters apart from each other. In case two or more images of the initial set are less than 2.5 meters apart, we select the most recent image within that group and filter out the rest. By applying this density filtering technique based on geographic location, we end up with our final set of 771,299 panoramic images. Each image is accompanied by metadata on geographic location, ground surface (either land or water), date and time taken, heading angle of the vehicle, and camera roll and pitch. The geo-localization error of a panorama is 0.02-0.08m under normal conditions, and up to 0.2-0.8m on rare occasions. The error margin on the viewing direction is at most 0.08\textdegree. Images are re-scaled to $1400 \times 700$ pixels.

\subsubsection{Elevation map data} The elevation data of Amsterdam was acquired via the online, open-source Current Dutch Elevation map \cite{ahn}. The elevation map data contains raster data with a resolution of 0.5 meters, acquired from LiDAR data with a point density of about 6 to 8 points per square meter. The elevation map was obtained by converting the point cloud to a 0.5 meter grid using a squared inverse distance weighting method \cite{ahn}. We use two elevation maps, the Digital Terrain Model (based on points marked as surface terrain), and the Digital Surface Model (based on points related to surface terrain and objects). The Digital Surface Model is subtracted from the Digital Terrain Model in order to acquire height information on large objects, such as buildings.

\subsubsection{Geospatial object information} We collected geospatial information on 22 different types of urban objects within Amsterdam by utilizing the data portal of the City of Amsterdam \cite{Data_ams} and OSM \cite{OSM}. The 22 classes included in the PanorAMS-noisy dataset concern the following objects that are very common in street scenery: \textit{building, tree, lamppost, bicycle path, park, bridge, waterway, bus, traffic sign, railway track, trash container, tram, train, public transport stop, playground, sport facility, windturbine, advertising column, high voltage pylon, traffic light, public toilet and ferry}.

\subsection{Automatically generating bounding boxes}

In this section, we describe our method for generating bounding box annotations in an automatic fashion (see Fig.~\ref{fig:automatic_labeling} for an overview). %

\subsubsection{Obtaining 3D real-world object coordinates}
The first step is to extract the relevant real-world 3D object measurements (e.g. the object's width, height, distance from the camera, and azimuth angle) from the multimedia sources (see step 1 in Fig. \ref{fig:automatic_labeling}). To this end, we first query the geospatial information on all objects falling within a 150 meter radius around the registered GPS location of the image. The 150 meter radius is a rough estimate of the maximum distance at which large objects may still be visible within our urban setting. For each relevant object we then determine the minimum distance between the object and the camera location. In case the location information is given by a polygon or a polyline, we are further able to determine the real-world width of the object as seen from the camera position. For polygons (e.g. buildings) we do this by first determining those points on the polygon for which a line can be drawn between the point and the camera location that does not cross the polygon at any other point. In case there are more than two points, we then determine which two of those points result in the largest angle between two lines, each drawn from one point on the polygon to the camera location. Once we have found two points, we approximate the real-world width of the object as the distance between the two points. When dealing with polylines (e.g. Railway Tracks), we first extract that part of the polyline that falls within the 150 meter radius area around the camera location. We then approximate the real-world width of the object as the distance between the two endpoints of the extracted part of the polyline. In case an object's location information does not allow us to determine the real-world width of the object, for instance if only a registered point location is available, the width is fixed to an estimated average of the object class (e.g. 1 meter for a lamppost). Here we can rely on prior observations of the urban space to estimate the real-world width of objects commonly encountered in street scenery, which for urban objects is a straightforward task to do as both the number of classes and the within-class visual appearance is generally limited.
We also need to determine the real-world height of the object. In case of relatively large objects, such as buildings or trees, we can determine the height based on elevation map data. For smaller objects, we again rely on our observations of the urban space to estimate the average height of a specific class of object (e.g. 6 meters for a lamppost). Finally, we determine the azimuth as the angle between the geographic north and the object's center point as seen from the camera location.
Following this procedure, we now have the necessary 3D world coordinates of each relevant object, namely the real-world height, width, minimum distance to the camera, and the azimuth angle.

\subsubsection{Camera model}
The second step is to map the 3D world coordinates of the queried objects to 2D image coordinates (see step 2 in Fig. \ref{fig:automatic_labeling}). Similar to  \cite{gis_eccv14}, we use the pinhole camera model to find an approximation for these coordinates. 
To apply the pinhole camera model, we first determine the camera model in the form of: $P = K[R|t]$, where $K$ is the intrinsic camera model, $R$ is the rotation matrix, and $t$ is a translation vector. $K$ can be obtained from: 

\[
K =
\begin{bmatrix}
s_{x} f & a & u_{0} & 0\\
0 & s_{y} f & v_{0} & 0 \\
0 & 0 & 1 & 0
\end{bmatrix}
\]

where $f$ is the focal length of the camera, $s_{x}$ and $s_{y}$ are the sensor size, $u_{0}$ and $v_{0}$ are the pixel coordinates of the image center, and $a$ is a skew factor related to the shear of the coordinate system. In our case, the images have already been calibrated for shear and rotation, leading to 0 shear and rotation in the formula above. $f$, $s_{x}$, $s_{y}$ are known from the camera specifications. $t$ includes a translation factor to account for the height of the camera, which is fixed based on camera specifications to 2 meters for images taken on land surface and 1 meter on water surface. 
We can now map the 3D real-world object coordinates ($x_r$, $y_r$, and $z_r$) to 2D image coordinates ($x_i$ and $y_i$) following: 
\[
\begin{bmatrix} 
x_i \\ 
y_i \\ 
1
\end{bmatrix}
= P
\begin{bmatrix}
x_r \\
y_r  \\
z_r 
\end{bmatrix}
\]

We apply the camera model two times to determine the position of the bounding box as though the most left-side of the object faces the true north. Both times $z_r$ concerns the distance between object and camera. The top-left pixel coordinates are then obtained by filling in 0 for $x_r$, and the real-world height of the object for $y_r$. The bottom-right coordinates follow from filling in the object's real-world width for $x_r$, and 0 for $y_r$. Based on the azimuth angle of the object's center point, we then determine the object's $x_{center}$ image coordinate and adjust the bounding box x-coordinates accordingly. By mapping the 3D coordinates of all queried objects to 2D image coordinates, we acquire an initial set of noisy bounding boxes.

\subsubsection{Box refinement and occlusion handling}

The initial set of bounding boxes still contains many boxes of objects that are occluded by other objects and, thus, needs to be refined and filtered (see steps 3-5 in Fig.~\ref{fig:automatic_labeling}). As buildings are large prominent objects, we address overlapping building boxes first. If a box overlaps entirely with another box of a building that is located closer to the camera, the occluded box further away from the camera is removed. In case a pair of boxes only partly overlap, the $x$ coordinates of the occluded box are adjusted such that there is no more overlap with the building box in the front. Next up, we observe that there are many places in the city where trees are packed closely together, resulting in extremely cluttered tree boxes in our initial set. To clean up this clutter, we remove tree boxes in case they overlap for more then 30\% with another tree box located closer to the camera. We then tackle the fact that there are overlapping boxes concerning the exact same object. This can happen when multiple sources of geospatial information are available for the same object class. In those cases, we merge overlapping boxes of the same class and make sure to retain the associated geospatial information of all sources. Finally, we jointly address all overlapping boxes of all classes. Boxes that overlap for more than 80\% with any box in the front, excluding boxes of classes that are considered non-blocking, are removed. Based on our city observations, the following classes are considered non-blocking; \emph{bicycle path}, \emph{railway track}, \emph{bridge}, \emph{park}, \emph{ferry}, \emph{bus}, \emph{waterway}, \emph{train}, \emph{tram}, \emph{tree}, and \emph{lamppost}.

\section{Efficient Crowdsourcing for Accurate Object Annotation: PanorAMS-clean Dataset}
In this section we describe how we selected the set of images contained in the PanorAMS-clean dataset, as well as our crowdsourcing protocol for obtaining the ground-truth bounding box annotations. We leverage Amazon Mechanical Turk \cite{MTurk} to select workers and conduct Human Intelligence Tasks (HITs).

\subsection{Image selection based on PanorAMS-noisy dataset}\label{sec:panorams:clean:image_select}

We construct the PanorAMS-clean dataset in such a way that it can be used 1) in conjunction with an overlapping subset of the PanorAMS-noisy dataset, e.g. to train teacher-student networks, 2) to evaluate performance of networks trained on the non-overlapping subset of images of the PanorAMS-noisy dataset, and 3) as a standalone dataset with clean annotations for fully supervised classification and detection. As such, we maintain three criteria for determining the subset of images of the PanorAMS-noisy dataset that form the PanorAMS-clean dataset. In order to prevent the over-representation of a geographic area, selecting a geographically balanced set of images is the first criterion. At the same time, following the reasoning behind group cross-validation commonly applied in e.g. medical image analysis and emotion recognition, all images of a certain administrative neighbourhood need to be assigned to either the PanorAMS-clean dataset or the train set of images of the PanorAMS-noisy dataset used in 2) as described above. This second criterion is important to guarantee sufficient disparity between the images contained in the test set and those in the train set. As a final criterion, all 22 classes of the PanorAMS-noisy dataset need to be present in PanorAMS-clean as well. In order to fulfill these criteria, we first determine based on PanorAMS-noisy which neighbourhoods contain objects belonging to rare classes. For each of the identified neighbourhoods we then inspect the geographic location and the variety of object classes over all images in the area. Based on this inspection and taking our crowdsourcing budget constraint for acquiring ground-truth annotations into consideration, we establish the PanorAMS-clean dataset of 7,348 images. The dataset covers 10 administrative neighbourhoods with a total land area of 3.9 km$^2$. %

\subsection{Annotation protocol}

\begin{figure}
  \centering
  \fbox{\includegraphics[width=0.5\textwidth]{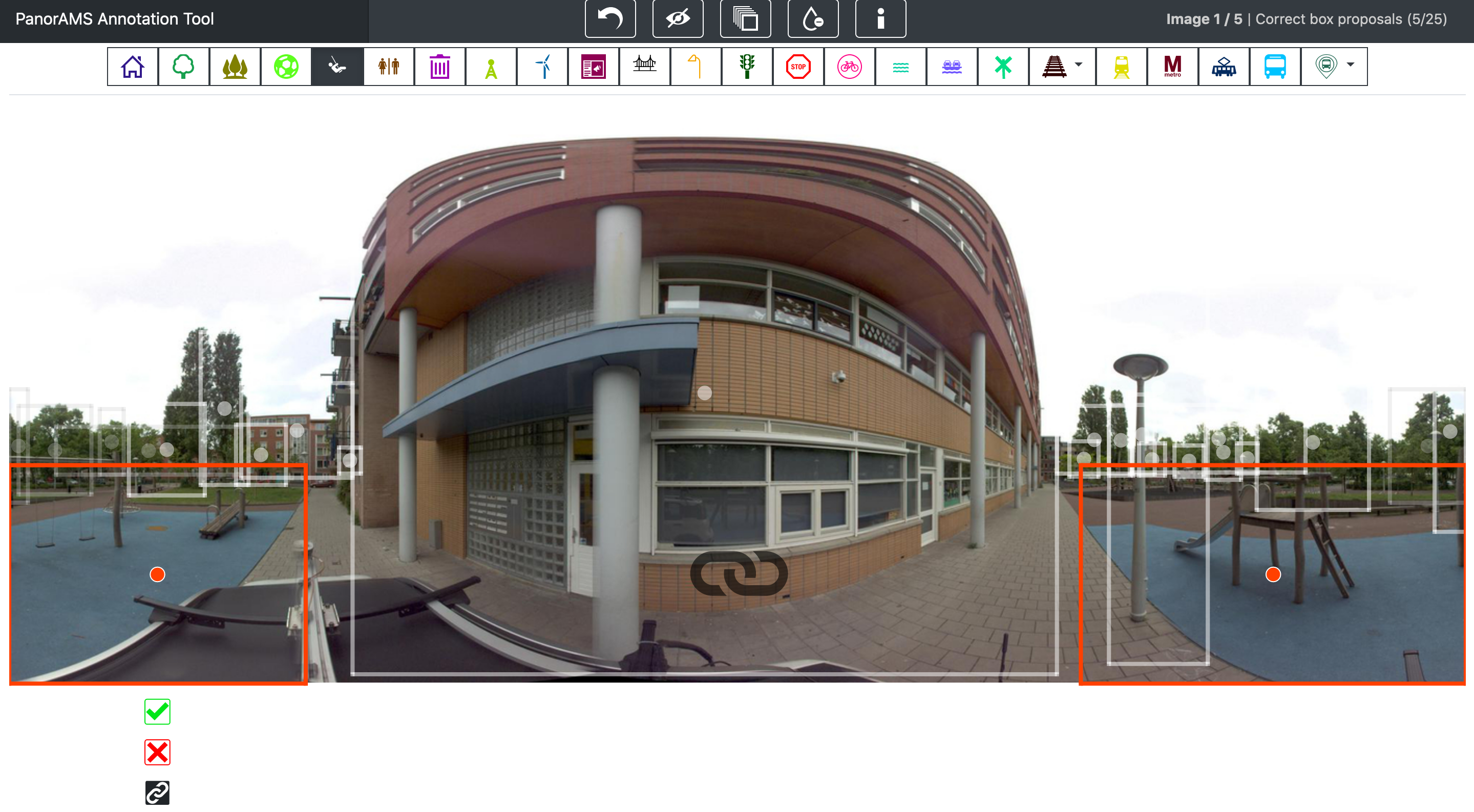}}
  \caption{The user interface of our crowdsourcing tool. Two linked boxes are active, ready to be broken up (by clicking the linkage button below the left active box), corrected (by dragging the middle point and borders of the active box) or deleted (by clicking the red X mark button) as need be. The linkage icon in the middle of the screen informs the user that there are two active linked boxes. The boxes can be verified by clicking the green check mark button. The orange color is specific for the playground class.}
  \label{fig:crowdsourcing_tool}
\end{figure}

In order to avoid task-switching, which is well-known to increase response time and decrease accuracy \cite{imagenet, crowdsourcing_annotations_object_detection}, our annotation protocol is subdivided into three tasks much like in \cite{crowdsourcing_annotations_object_detection}. 
One significant difference with \cite{crowdsourcing_annotations_object_detection} is that we do not need to start from scratch. We already have our automatically generated bounding boxes following the approach presented in Section~\ref{sec:panorams:noisy}. 
Having initial boxes to start with helps workers localize objects, especially in crowded street scenes. We further introduce a specific labeling method for objects that span the left- and right-side of the image (due to the nature of 360° panoramic images), whereby two boxes need to be linked. Linked boxes are fixed to the right and left extremities of the image and have equal height (see Figure~\ref{fig:crowdsourcing_tool} for an example). 

\begin{figure*}
  \centering
  \includegraphics[width=\textwidth]{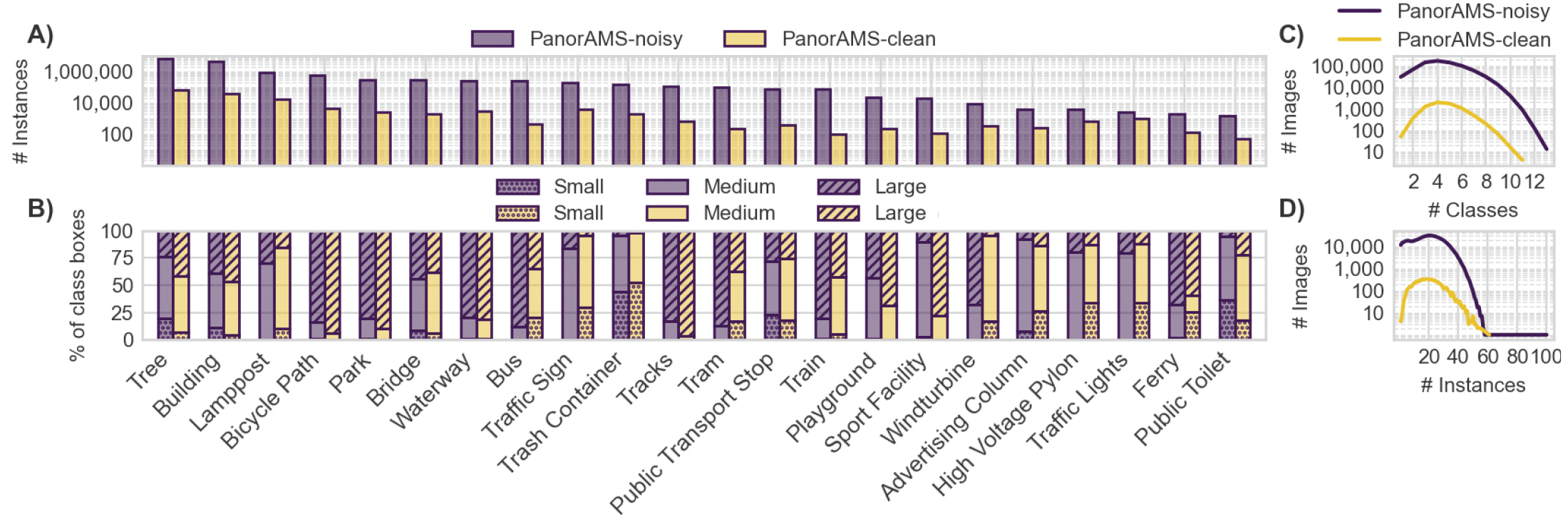}
  \caption{A) Number of instances per class, B) percentage of small, medium, and large bounding boxes per class, C) number of classes per image, and D) number of instances per image in PanorAMS-noisy (purple) and PanorAMS-clean (yellow).}
    \label{fig:instances_dist}
\end{figure*}

\subsubsection{Adjustment task}
In this task a worker corrects, with the possibility to delete, the bounding boxes that have been automatically generated (see Section~\ref{sec:panorams:noisy}). This is done for all classes simultaneously \cite{ferrari2019}. The annotator's attention is directed from left to right by starting the correction with the utmost left box and then continuing the correction with each subsequent box to the right. The current box under consideration is indicated by a specific color corresponding to the object class the box belongs to. Upon hovering over the box an icon is shown, which also corresponds to the box' object class. 

\subsubsection{Addition task}

The main purpose of this task is to collect bounding boxes for those objects that were not included during the first task. In order to do so, workers need to know which boxes have already been labelled in an image. Thus, per class we first have a verification stage in which workers verify and, if necessary, further adjust each box that was acquired during task 1. This way we provide workers the necessary information concerning which objects have already been labeled, while at the same time building in a quality check on the already labeled boxes. After each class verification stage, a worker can add new boxes for instances of the specific class that were missed during task 1. Workers can add new boxes by clicking on the four extreme points of the object, i.e. top, bottom, left-most, right-most. This method, called extreme clicking, has been shown to significantly reduce annotation time compared to drawing a box in the conventional manner \cite{extreme_clicking}.

\subsubsection{Final verification task}
The last task serves as a final verification and quality check. It follows the same setup as the addition task. Per class, workers are guided through the image from left to right by having to first verify the boxes acquired during task 2. After each class verification stage, they can then add new boxes for the specific class by means of extreme clicking.

\subsection{Annotator evaluation and training}
Workers are first presented with a step-by-step tutorial explaining the annotation process and functionality of the tool. Per class we explain what the correct way of labeling an instance is, accompanied by examples of well-labelled bounding boxes. After the tutorial, workers need to pass a qualification test before they can start the official task. In case of failure, workers can review their work and compare it to the bounding boxes used as evaluation. This way workers get immediate feedback on their work, allowing them to learn from their mistakes. Workers can retake the qualification test as many times as needed.

Similarly to \cite{extreme_clicking}, we provide workers with a batch of 5 images. Each batch contains 1 gold standard image that is used to evaluate workers' performance on the HIT. Gold standard images are acquired by selecting a geographically balanced subset of images of the PanorAMS-clean dataset (482 images) and recruiting multiple well-informed computer vision experts to label those images. A final verification check on all gold standards images by one and the same expert safeguards the quality of gold standard annotations. We set an absolute minimum quality threshold on gold standard images in order to automatically filter out poor quality work. Work that does not pass this quality threshold is automatically rejected. We perform an additional, more extensive evaluation on work that did pass the automatic evaluation stage.

\begin{figure*}
  \centering
  \includegraphics[width=\textwidth]{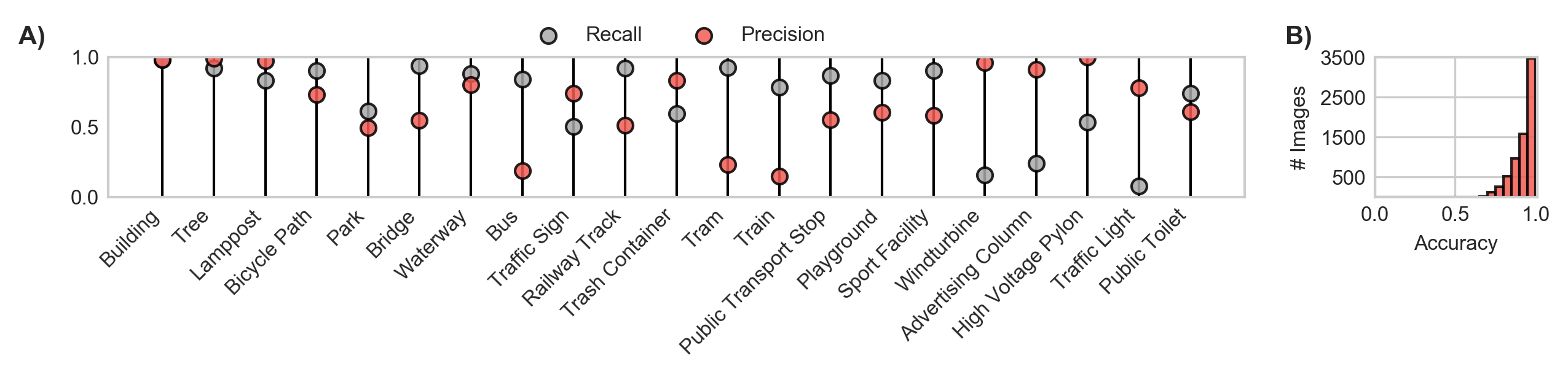}
  \caption{A) Precision and recall statistics per class, and B) overall image label accuracy of comparing a subset of the PanorAMS-noisy image class labels with the labels of the same set of images of PanorAMS-clean. 
  Classes on the x-axis in A) are ordered starting with the class having the highest amount of positive image labels in PanorAMS-noisy on the left to the lowest on the right.}
    \label{fig:img_label_noise_stats}
\end{figure*}

\begin{figure*}
  \centering
  \includegraphics[width=\textwidth]{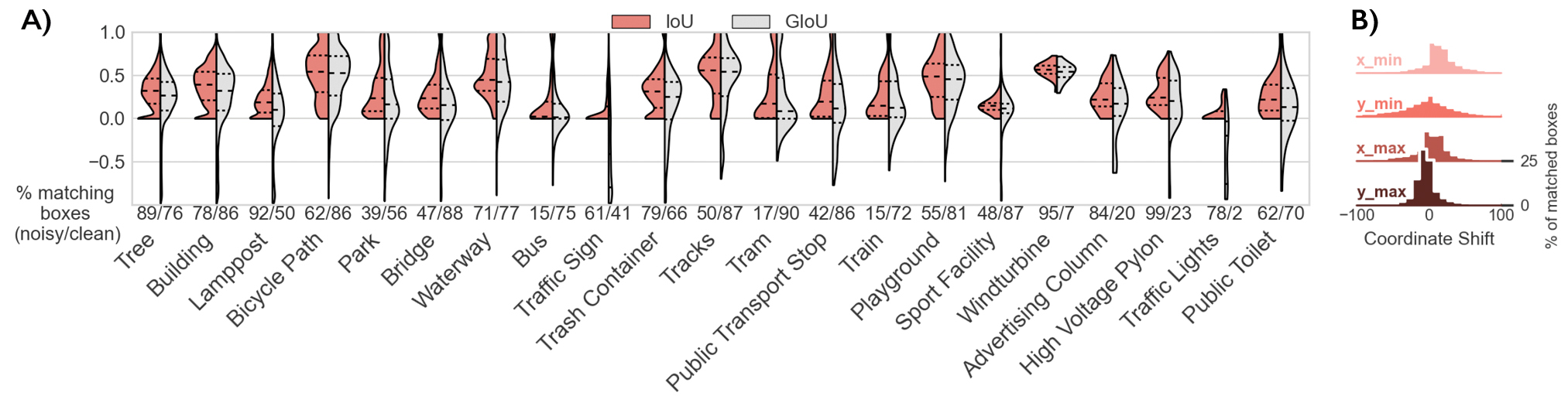}
  \caption{A) Overlap statistics per class, and B) overall coordinate difference between matched boxes of PanorAMS-clean and the overlapping subset of PanorAMS-noisy. Coordinate shift values were obtained by subtracting the bounding box coordinates from PanorAMS-clean from the bounding box coordinates of matched boxes of PanorAMS-noisy. Classes on the x-axis in A) are ordered starting with the class having the highest number of bounding box instances on the left to the lowest on the right.}
    \label{fig:noise_rate_stats}
\end{figure*}

\subsection{Statistics on annotation results}

To validate the quality of our crowdsourcing effort, we analyze the median Intersection over Union (IoU) obtained on gold standard images. For the same expert, redoing the labeling of gold standard images resulted in an average of 0.83 median IoU. This highlights the difficulty of labeling crowded street scenes in which there are many small and partially occluded objects (see Figures~\ref{fig:instances_dist} and ~\ref{fig:detection_qualitative}). On average, MTurk workers obtained a median IoU of 0.72 with gold standard images. Further crowdsourcing statistics per administrative neighbourhood can be found in the supplementary material.

During the first phase, the median time spent on adjusting a box is 8.9 seconds, including the action of deleting boxes. If we ignore deleting boxes as an action (which only takes a median 2.4 seconds), the adjustment time per box increases to 10.3 seconds. Interestingly, this is a lot faster than the reported 34.5 seconds for drawing and verifying a box in \cite{crowdsourcing_annotations_object_detection} and is only slightly slower than the reported 7.0 seconds for creating a new box with extreme clicking in \cite{extreme_clicking}. One possible explanation for our fast annotation time, is that using the generated boxes as a starting point reduces the time workers need to spend on visually searching for objects in the image. Contributing factors may be that we kept required mouse movements to a minimum and allowed for easy adjustment of boxes. During the second task, the median time spent on verifying and adding new boxes is 5.9 seconds per box. During the third task, this median time drops to 4.4 seconds per box as the quality of the annotations increases and, thus, the time required to correct a box decreases.

\section{Dataset Statistics}

In this section we provide key insights into the statistics of the PanorAMS-noisy and PanorAMS-clean datasets.

\subsection{Distribution of Instances and Classes}\label{sec:dataset_stats:dist_instances_classes}

Figure~\ref{fig:instances_dist}A shows the total number of instances per class in PanorAMS-noisy and PanorAMS-clean. It is apparent that both datasets involve large class imbalances. Figure~\ref{fig:instances_dist}B further specifies for each class what percentage of the bounding boxes are small, medium and large in size. We follow the definition originally outlined in MS COCO that bounding boxes with an area < 32\textsuperscript{2} are small,  32\textsuperscript{2} <= area <= 96\textsuperscript{2} are medium, and area > 96\textsuperscript{2} are large \cite{ms_coco}. There is significant variation among the different classes as to the size of objects, whereby PanorAMS-noisy and PanorAMS-clean follow a similar pattern for the majority of the classes. Figure~\ref{fig:instances_dist}C shows the distribution of unique classes in images, and figure~\ref{fig:instances_dist}D shows the distribution of the total number of instances in images. Both datasets clearly involve crowded scenes with many instances and multiple different classes per image.

\subsection{Noise Analysis}

In order to estimate the amount of image label noise in PanorAMS-noisy, we analyse the precision and recall per class, and the overall label accuracy per image by comparing a subset of the image labels of PanorAMS-noisy with the labels of the same set of images of PanorAMS-clean (see Fig.~\ref{fig:img_label_noise_stats}). The three most prevalent classes (Building, Tree, and Lamppost) all have high precision and recall (> 0.85). The dynamic classes (Bus, Tram, and Train) have, as expected, low precision (< 0.25), while their recall is quite high (> 0.75). The latter is due to the fact that these types of objects generally follow a clear, accurately defined geographic route. A few classes (Windturbine, Advertising Column, and Traffic Light) have low recall (< 0.25), while their precision is quite high (> 0.75). Overall, the label accuracy per image is > 0.6, with over 70\% of the samples > 0.9.

As estimation for the amount of bounding box location noise in PanorAMS-noisy, we analyse the overlap between bounding boxes of PanorAMS-clean and the bounding boxes of the overlapping subset of images of PanorAMS-noisy. We use two overlap metrics, namely Intersection over Union (IoU) and Generalized IoU (GIoU) \cite{GIoU_2019_CVPR}. IoU, the most commonly used metric for evaluating the similarity between two bounding boxes (A, B), is defined as \(IoU = \frac{A \cap B}{A \cup B}\). One major drawback of IoU is that if \(A \cap B=0\), then \(IoU(A,B)=0\), no matter how far or close the bounding boxes are to each other. GIoU aims to solve this issue by adding a penalty term as follows \(GIoU = IoU - \frac{|C\setminus(A \cap B)|}{|C|}\), whereby \(C\) is the smallest box enclosing \(A\) and \(B\). Using these metrics, we solve the optimal assignment between noisy and clean bounding boxes with the Hungarian algorithm \cite{Kuhn1955Hungarian}. Figure~\ref{fig:noise_rate_stats}A shows the result of the matching. For each class the distribution of IoU and GIoU values is given, as well as the percentage of noisy and clean boxes that could be matched. There are clear class differences. For some classes, such as tree and building, a high percentage of the noisy boxes can be matched with a high percentage of the clean boxes and at least half of the boxes have an IoU/GIoU > 0.3. Bicycle path and railway tracks have even higher IoU/GIoU values for most boxes, but a relatively higher percentage of noisy boxes that could not be matched with any clean box. There are also extremely noisy classes, both in terms of IoU/GIoU values and the percentage of boxes that could be matched, such as traffic sign and traffic light. 
We further analyse the overall shift from clean to noisy bounding box coordinates. For this analysis, we calculate the Euclidean distance between noisy and clean x\textsubscript{min}, y\textsubscript{min}, x\textsubscript{max}, and y\textsubscript{max} coordinates without differentiating between classes, and again use the Hungarian algorithm to match boxes. Figure~\ref{fig:noise_rate_stats}B shows the result. Generally, the noisy boxes are slightly shifted to the right compared to clean boxes. The y\textsubscript{max} coordinate is relatively accurate in most cases, while there is a larger variance in the error on the y\textsubscript{min} coordinate.

\section{Experimental Setup}
In this section we describe the experimental setup for evaluating classification and detection performance on our PanorAMS datasets. Unless stated otherwise, mentioned parameter settings were found via grid search using the train-val split of the PanorAMS-clean dataset (when training on clean annotations) or the train-val split of the PanorAMS-noisy dataset (when training on noisy annotations).

\subsection{Data}
\label{sec:exp:data}
In order to improve performance for objects that are split across the left and right extremities of the panoramic image, we circularly pad the left-side of the image with 25 pixels of the rightmost part of the image and vice-versa for the right-side of the image. Bounding boxes are corrected for this increase in image size by increasing the width of linked boxes accordingly based on the amount of width of the object present in the circularly padded area and duplicating (part of) the non-linked bounding boxes that have a minimum width of 20 pixels in the padded area. The bottom 150 pixels of the image are cropped off as this part solely contains the capturing vehicle. This results in the final detection input size of $1450 \times 550$ pixels. For classification, the top 50 pixels, which contain heavy object distortions due to the panoramic nature, are cropped off. The remaining image of $1450 \times 500$ pixels is cut into three pieces of each $500 \times 500$ pixels, whereby each image part has an overlay of 25 pixels with each of the other two parts. Each of these parts is resized to the classification input size of $224 \times 224$ pixels.

\subsubsection{Augmentation} 

We apply random augmentations during train time in order to regularize training and prevent
over-fitting. During classification, we apply mixup ($\alpha$ = 0.2) \cite{mixup} and random image resizing in the range of $180 \times 180$ and $224 \times 224$ pixels while maintaining the original aspect ratio of the image. During detection, we apply the mixup augmentation as implemented in the mmdetection toolbox \cite{mmdetection}, random image resizing in the range of $1160 \times 440$ and $1450 \times 550$ pixels while keeping the aspect ratio intact, and random cropping whereby the cropped image and bounding boxes have a minimum IoU of 0.5 with the original.

\subsubsection{Oversampling}

To counter the large class imbalance in our datasets, we train all models with repeat factor sampling (RFS) \cite{gupta_lvis_2019}, which has been shown to yield state-of-the-art results on class imbalanced datasets. For each category \(c\), the category-level repeat factor is defined following \(
r_{c} = max(1, \sqrt{t/f_{c}})
\), whereby \(f_{c}\) is the fraction of images that contain at least one instance of category \(c\) and \(t\) is a hyperparameter. Since an image may contain multiple categories, the image-level repeat factor is then defined following \(r_{i} = max_{c\in i}(r_{c})\), whereby \(\{c\in i\}\) are the categories labeled in image \(i\). Intuitively, this means that all images containing a category with \(f_{c}\) > \(t\) are oversampled following the square-root inverse frequency heuristic of the rarest class present in the image. We empirically found \(t = 0.1\) to work best in our setting.

\begin{table}
\caption{PanorAMS Dataset Splits}
\label{tab:datasplits}
\centering
\begin{tabular}{llcr}
\toprule
   \rule{0pt}{1.1em}  & Dataset Split & Annotations & {\# Images} \\ [0.5em]
\midrule
\multirow{3}{*}{\rotatebox[origin=c]{90}{\parbox[c]{1cm}{\centering Training}}}
\rule{0pt}{1.1em} & Clean (subset) & PanorAMS-clean & 5,817 \\
 & Noisy (subset) & PanorAMS-noisy & 5,817 \\
 & Noisy (all) & PanorAMS-noisy & 769,437 \\ [0.5em]
\midrule
\multirow{1}{*}{\rotatebox[origin=c]{90}{\parbox[c]{0.41cm}{\centering Test}}}
\rule{0pt}{1.1em} & Clean (test) & PanorAMS-clean & 1,444 \\ [0.5em]
\bottomrule
\end{tabular}
\end{table}

\subsubsection{Splits} 
We split the data following the criterion outlined in 
Section~\ref{sec:panorams:clean:image_select} that all images of a certain administrative neighbourhood need to be assigned to either the train, validation or test split. This results in 4,213 and 763,951 train, 1,606 and 5,657 validation, and 1,444 and 1,444 test images in, respectively, the PanorAMS-clean and PanorAMS-noisy dataset. Additionally, we create a subset dataset of PanorAMS-noisy, containing the same images, and splits as the PanorAMS-clean dataset. Due to splitting according to administrative neighbourhoods, it is not possible to select sufficient instances for all classes in each split. We've had to exclude Wind Turbine, High Voltage Pylon, Sport Facility, and Train from our evaluation as there are no or too few examples (< 6) in the Clean (test) split, and Ferry as there are no training examples in the Noisy (subset) split. Parameter settings were optimized on the validation set, after which the train and validation split were used as the training set for all final experiments. The dataset splits used for the final experiments are shown in Table \ref{tab:datasplits}. Clean (subset) and Noisy (subset) concern the same set of images. All models are evaluated on the Clean (test) split.

\subsection{Classification} 
To evaluate classification performance on our PanorAMS datasets, we use the popular ResNet-50 model \cite{resnet50}. The model is trained from scratch in a multi-label fashion, whereby the three image parts (acquired by following the procedure described in Section~\ref{sec:exp:data}) are processed by the network in a round robin fashion, i.e. the first image part goes through the network first, the second part second, and the third part last. Network weights are updated after all three parts of the image have been processed.

We use the F-score as evaluation metric, which is defined as \(F = 2* \frac{precision * recall}{precision+recall}\). We use the scikit-learn implementation of the weighted F-score as aggregate for all classes. This is the macro F-score weighted according to the number of true instances for each class in order to account for class imbalance.

\subsection{Detection} 
To evaluate detection performance on our PanorAMS datasets, we use two common general detectors, the two-stage detector Mask R-CNN \cite{mask_rcnn} and the cascade detector Cascade R-CNN \cite{cascade_rcnn}. We favor general detectors over more domain specific networks such as pedestrian detectors. As an example consider pedestrian detectors, they are specifically developed to tackle dense object detection, but a recent study found such models to poorly handle even small domain shifts \cite{elephant_cvpr_2021}. Cascade R-CNN was found to generalize well to pedestrian detection, achieving competitive results compared to recent state-of-the-art pedestrian detectors. We use a ResNet-50 backbone pretrained on ImageNet with 6 level FPN \cite{fpn} for both Mask R-CNN and Cascade R-CNN. The pretrained weights of the first two layers of the backbone are frozen. We adjust the anchor generation settings to account for the great variety in object scales and sizes present in the PanorAMS datasets. We set the anchor ratios to [0.15, 0.6, 1.3, 3.5, 4.5, 7.5, 11.5] based on the median bounding box width:height ratios per class in PanorAMS-noisy. Anchor strides are set to [4, 8, 16, 32, 64, 128] based on the bounding box sizes per class in PanorAMS-noisy. In the clean annotation setting, the background and the foreground thresholds to determining negative and positive samples are set to, respectively, 0.25 and 0.7 for both models. In the noisy annotation setting, the background and foreground thresholds are dropped to, respectively, 0.1 and 0.6 in order to account for the bounding box location noise. In the cascade model, the foreground thresholds of the cascade detectors are set to 0.4, 0.5, and 0.6 in the noisy setting. We empirically found these parameters to work best on relevant validation sets. Further implementation details for Mask R-CNN and Cascade R-CNN are based on the default settings of the mmdetection toolbox \cite{mmdetection}.

As evaluation metric, we use the COCO-style average precision (mAP@[.5;.95] ) metric that averages over categories and different box IoU thresholds \cite{ms_coco}, mAP@[.5] that measures average precision at an IoU threshold of 0.5, and recall@100 that measures recall when only the top 100 detections are selected from an image.

\subsection{Training Details}

We optimize all networks using the stochastic  gradient descent (SGD) algorithm with gradient centralization, which has recently been shown to stabilize training and improve generalization \cite{gradient_cent}. Weight decay and momentum are fixed to, respectively, 0.0001 and 0.9 for all experiments. As DNNs learn simple patterns first, it should be beneficial to show simpler images with fewer objects first. We therefore train on images in an ordered fashion whereby the image containing the least amount of object instances is seen first and the image containing the largest amount of object instances last. We indeed empirically verified this to be the most stable way of training. In order to prevent overfitting on the Noisy (all) split, we divided the examples of this split into 120 different sets, whereby each set is geographically balanced and all images in a set are ordered from least amount of total object instances to largest. We train all networks using the 1cycle learning rate policy \cite{smith2017superconvergence} with a batch size of 5, initial learning rate of 0.002, linear warm-up of 500 iterations (0.001 warm-up ratio), and a maximum learning rate of 0.02. We train for 120 (= $1\times$ the entire dataset split), 12, and 12 epochs on, respectively, the Noisy (all), Noisy (subset), and Clean (subset) split.

\section{Results}
In this section we discuss our classification and detection results. 

\subsection{Classification}

\begin{figure}
  \centering
  \includegraphics[width=0.5\textwidth]{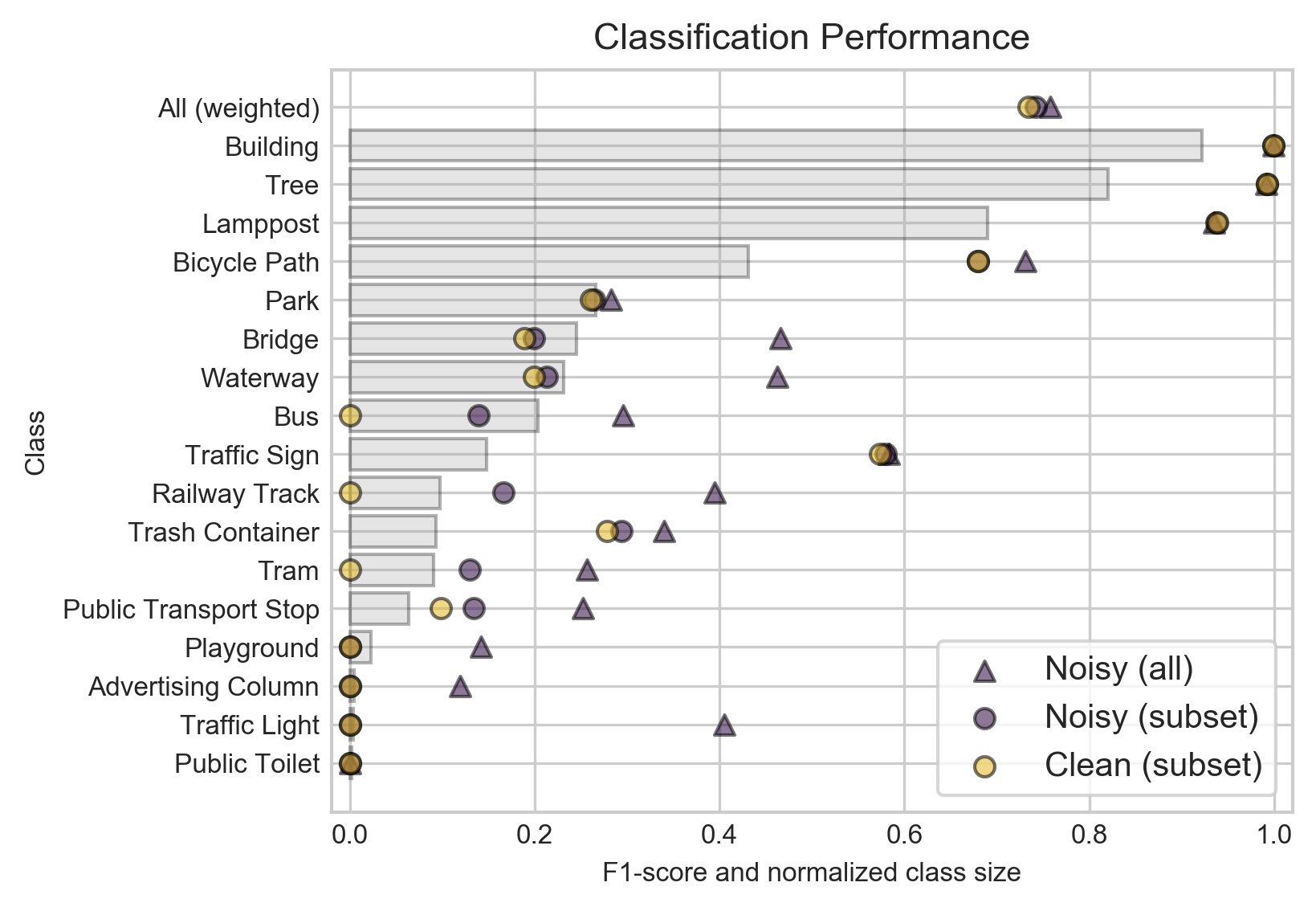}
  \caption{Overall and per class classification performance of ResNet-50 model trained on different training splits of the PanorAMS datasets. Models are evaluated on the same test set. Grey bars depict normalized class size based on all samples in the PanorAMS-noisy dataset.}
   \label{fig:classification}
\end{figure}
\begin{figure*}
  \centering
  \includegraphics[width=\textwidth]{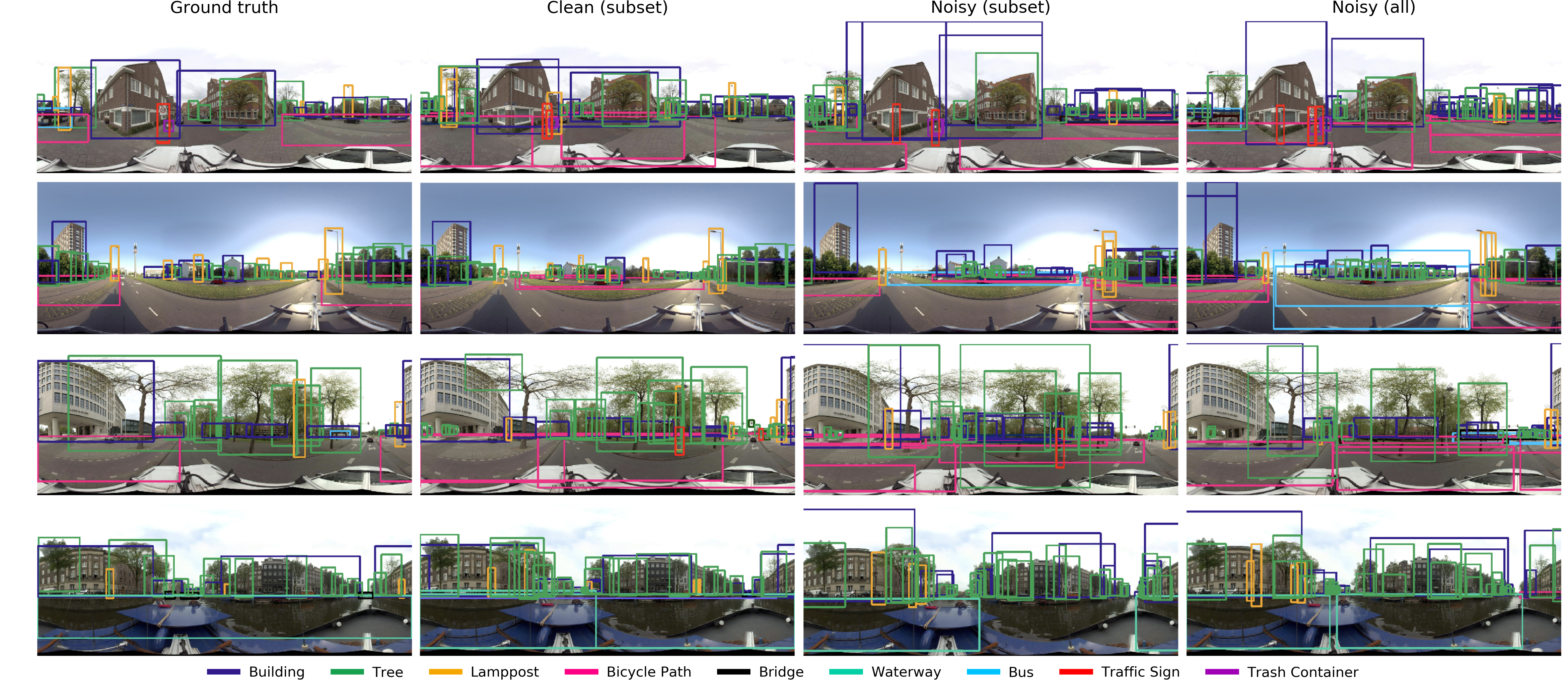}
  \caption{Qualitative results of Cascade R-CNN trained on different training splits of the PanorAMS datasets.}
   \label{fig:detection_qualitative}
\end{figure*}

Figure \ref{fig:classification} summarizes the overall and per class classification performance on the PanorAMS datasets. We see high performance overall, but there are clear performance differences per class. The three most prevalent classes (Building, Tree, and Lamppost) all show comparable performance on each training split. We observe that for classes with fewer training examples training on the Noisy (all) dataset results in better performance than training on the much smaller Clean (subset) dataset. The large increase in training examples outweighs the negative effect of training with noisy image labels. Notably, especially polyline objects (e.g. Bridge, Waterway, and Railway Track) appear to profit most from the increase in training examples in the Noisy (all) dataset. The geographic location of these objects is generally well-defined and accurate as this is concerning mostly rigid objects, resulting in relatively clean (high recall) image labels for those classes (see Figure~\ref{fig:img_label_noise_stats}A). Overall, training on the much larger Noisy (all) training split leads to similar or higher performance for all classes compared to training on a much smaller training split with accurate image labels. Our estimated label accuracy per image of the PanorAMS-noisy dataset is quite high (see Figure~\ref{fig:img_label_noise_stats}B), which could explain our high performance on the Noisy (all) split.
Figure \ref{fig:classification} further reveals the difficulty of obtaining (good) classification performance on classes belonging to the long-tail of infrequent classes. Further techniques to deal with the heavy class imbalance may improve performance for these tail classes. We provide additional precision-recall curves per class in the supplement.

\subsection{Detection}

\begin{table}
\caption{Detection Performance}
\label{tab:detection}
\centering
\begin{tabular}{llccc}
\toprule
   \rule{0pt}{1.1em}  & Trainset & mAP@[.5;.95] & mAP@.5 & recall@100 \\ [0.5em]
\midrule
\multirow{3}{*}{\rotatebox[origin=c]{90}{\parbox[c]{1cm}{\centering Mask R-CNN}}}
\rule{0pt}{1.1em} & Clean (subset) &             \textbf{10.2} &   \textbf{23.3} &       \textbf{20.2} \\
& Noisy (subset) &     1.5 &    5.9 &        4.9 \\
& Noisy (all) &      2.5 &    9.2 &        7.5 \\ [0.5em]
\midrule
\multirow{3}{*}{\rotatebox[origin=c]{90}{\parbox[c]{1cm}{\centering Cascade R-CNN}}}
\rule{0pt}{1.1em} & Clean (subset) &    9.8 &   21.6 &       19.3 \\
& Noisy (subset) &  1.2 &    5.2 &        5.1 \\
& Noisy (all) &   2.6 &    9.2 &        8.1 \\ [0.5em]
\bottomrule
\end{tabular}
\end{table}

\begin{table*} 
\caption{Detection Performance on the PanorAMS datasets, mAP@.5}
\label{tab:detection_per_topic}
\centering
\begin{tabular}{llccccccccc}
\toprule
     \rule{0pt}{1.1em}  & Trainset & Building &  Tree & Lamppost & Bicycle Path & Bridge & Waterway & Traffic Sign & Railway Track & Trash Container \\
\midrule
 \multirow{3}{*}{\rotatebox[origin=c]{90}{\parbox[c]{1cm}{\centering Mask R-CNN}}}
\rule{0pt}{1.1em} & Clean (subset) & \textbf{66.5} &  \textbf{71.6} &     \textbf{57.0} &         19.2 &    \textbf{9.3} &     36.5 &         29.5 &          16.4 &            \textbf{35.2} \\
& Noisy (subset) & 27.1 &  24.9 &      4.5 &          8.3 &    0.7 &     25.5 &          0.2 &           3.9 &             4.5 \\
& Noisy (all) &   38.9 &  26.7 &      3.0 &         21.4 &    2.5 &     40.6 &          0.7 &          15.0 &             5.9 \\
\midrule
 \multirow{3}{*}{\rotatebox[origin=c]{90}{\parbox[c]{1cm}{\centering Cascade R-CNN}}}
\rule{0pt}{1.1em} & Clean (subset) &     65.2 &  69.8 &     53.8 &         19.6 &    4.1 &     38.4 &         \textbf{30.1} &          19.2 &            33.8 \\
 & Noisy (subset) &     28.6 &  21.9 &      2.9 &          8.4 &    1.5 &     19.5 &          0.2 &           0.1 &             4.9 \\
   & Noisy (all) &     36.5 &  23.3 &      3.6 &         \textbf{23.7} &    0.8 &     \textbf{41.1} &          0.9 &     \textbf{20.0} &        6.2 \\
\bottomrule
\end{tabular}

\end{table*}

In Table~\ref{tab:detection} we provide the overall mAP@[.5;.95], mAP@.5 and recall@100 for the PanorAMS datasets. The low mAP@[.5;.95] score on the Clean (subset) trainset demonstrates that PanorAMS is a challenging detection dataset, even with accurate bounding box annotations. As illustrated in Figures~\ref{fig:instances_dist} and ~\ref{fig:detection_qualitative}, oftentimes images are crowded, containing box annotations of multiple object instances, different classes, and a variety of object sizes. The overall scores drop significantly when training on noisy annotations, highlighting the even bigger challenge of performing object detection in a noisy label setting. For 8 classes (Park, Bus, Tram, Public Transport Stop, Playground, Advertising Column, Traffic Light, and Public Toilet) the extent of the noise inhibits the network from learning to recognize and localize these objects in our supervised setting. Further research on specific noise handling techniques is required to reveal whether sufficient training signal can be obtained from our noisy annotations for these classes. Inspecting the detection performance on the other 9 classes (see Table~\ref{tab:detection_per_topic}), reveals interesting differences. In line with what was observed during classification, large, rigid, polyline objects (Waterway, Bicycle Path, and Railway Track) benefit from the increase in training examples in the Noisy (all) split and even outperform the network trained on the Clean (subset) split. In Figure~\ref{fig:noise_rate_stats}A we observe that the median IoU/GIoU for bounding boxes of these classes is around or above 0.5. The percentage of matching clean and noisy boxes is also relatively high with at least 50\% of the noisy bounding boxes overlapping with at least 50\% of the clean boxes. In addition, the majority of these boxes is large in size (see Figure~\ref{fig:instances_dist}B). Notably, for the polyline object class that does not perform well (Bridge) the overlap between noisy and clean boxes, and the percentage of large boxes are both significantly lower. Less than 25\% of the noisy bridge boxes have an IoU/GIoU of 0.5 or higher with clean boxes. This observation is true for all classes with mAP@.5 < 10\% on the Noisy (all) split (Lamppost, Bridge, Traffic Sign, and Trash Container). The highest performing of all classes with mAP@.5 < 10\% (Trash Container) also displays the most overlap between noisy and clean bounding boxes in comparison. Finally, the two most prevalent classes (Building and Tree) both show decent performance on the Noisy (all) split, while the performance gap with the Clean (subset) split is apparent. Despite a large number of different instances to learn from, the network clearly suffers from the noisy bounding box annotations. This becomes especially apparent when comparing the performance between the Noisy (subset) and Noisy (all) split. Generally, we observe that the noisier the annotations of a class are, the less that class is able to benefit from the increase in training examples in the Noisy (all) split. This highlights the limitations of training current fully supervised detection networks in the presence of considerable class and bounding box location noise.

\section{Conclusion}
We introduced the PanorAMS framework, which is designed to move away from the costly effort of manual annotation for object detection. By generating bounding boxes around a range of objects in public spaces solely based on urban context information and linking them to panoramic images, we open the door to automatically generating information-rich urban multimedia at an unprecedented scale that could be deployed in numerous tasks of relevance to the multimedia community. Together with the efficiently crowdsourced ground-truth annotations of PanorAMS-clean, our large-scale PanorAMS-noisy dataset enables future study of classification and object detection in a real-world setting with annotations involving both class and bounding box location noise. This is a challenging task, as our results reveal the detrimental effect bounding box location noise can have on detection performance as well as how different types of noise affect network performance differently. It may take a combination of techniques before networks are truly capable of differentiating noisy signals from general signals of interest about an object’s location. On the other hand, we already obtain improved classification performance training on our large set of noisy image labels compared to training on a much smaller set of clean labels. This demonstrates the capacity of DNNs to ignore noise to a large extent, provided that sufficient examples with sufficient training signal are present. Our analysis is an important step in determining where those thresholds regarding the amount of examples and training signal that constitute sufficient may lie.

\section*{Acknowledgment}
This work was funded in part by the European Regional Development Fund (ERDF) Interreg VB North Sea Region Programme project Smart Cities + Open Data Re-use (SCORE) and has received funding from the City of Amsterdam. The work is supported by the NVIDIA GPU Grant Program.

\ifCLASSOPTIONcaptionsoff
  \newpage
\fi

\bibliographystyle{IEEEtran}

\end{document}


\title{Supplementary Material for \\ PanorAMS: Automatic Annotation for \\ Detecting Objects in Urban Context}
\author{Inske Groenen,
        Stevan Rudinac,
        and~Marcel Worring,~\IEEEmembership{Senior Member,~IEEE}}

\hypersetup{pageanchor=false}
\maketitle
\hypersetup{pageanchor=true}

In this supplement, we provide additional information on:
\begin{itemize}
\item Object information from multimedia sources,
\item Crowdsourcing statistics,
\item Panoramic distortion,
\item Results.
\end{itemize}

\section{Object information from multimedia sources}

Table~\ref{tab:object_info} specifies the object information obtained from multimedia sources. As stated in Section III-B of the main paper, width and height estimates are based on city observations and are only used in case, respectively, width or height information can not be obtained from other multimedia sources. 
\begin{table}[h]
  \caption{Urban object information}
  \label{tab:object_info}
  {\small
\begin{tabular}{lrrrr}
\toprule
Class &  \thead{Data \\ Sources} &  \thead{ Width \\ Estimate } & \thead{Height \\ Estimate} \\
\midrule
Advertising Column                    &       Geospatial &                  1.3 &                 3.2 \\
Bicycle Path           &        Geospatial &                           4.0 &        1.5 \\
Building                     &        \makecell{Geospatial \& \\ Elevation Map} &                  5.0 &                 10.0 \\
Bus                  &       Geospatial &                  4.0 &                 2.5 \\
Bridge                   &       \makecell{Geospatial \& \\ Elevation Map} &                  4.0 &                 2.5  \\
Ferry                    &       Geospatial &                  4.0 &                 3.5 \\
High Voltage Pylon                    &       Geospatial &                  15.0 &                 25.0 \\
Lamppost          &        Geospatial &                  1.0 &                 6.0  \\
Park &       Geospatial &                 5.0 &                10.0  \\
Playground                    &       Geospatial &                  7.0 &                 2.0 \\
Public Toilet                    &       Geospatial &                  1.5 &                 2.5 \\
Public Transport Stop                    &       Geospatial &                  2.5 &                 2.0 \\
Railway Tracks                    &       Geospatial &                  4.0 &                 1.5 \\
Sport Facility                    &       Geospatial &                  10.0 &                 3.0 \\
Traffic Light                    &       Geospatial &                  0.5 &                 2.5 \\
Traffic Sign                    &       Geospatial &                  0.5 &                 2.5  \\
Train                    &       Geospatial &                  4.0 &                 4.0 \\
Tram                    &       Geospatial &                  4.0 &                 2.5  \\
Trash Container                  &       Geospatial &                  1.2 &                 1.5 \\
Tree                       &       \makecell{Geospatial \& \\ Elevation Map} &                 5.0 &      {-} \\
Waterway             &       Geospatial &                  4.0 &                 1.5  \\
Windturbine                    &       Geospatial &        30.0 &                50.0 \\
\bottomrule
\end{tabular}
  }
\end{table}

\section{Crowdsourcing statistics}

Table~\ref{tab:crowdsourcing_perf} summarizes crowdsourcing statistics per administrative region in PanorAMS-clean. As stated in Section IV-D, crowdsourcing is evaluated using IoU agreement between expert and MTurk worker annotations on gold standard images.

\begin{table*}
  \caption{Crowdsourcing statistics}
  \label{tab:crowdsourcing_perf}
  {\small
\begin{tabular}{lrrrrrr}
\toprule
{Administrative Neighbourhood} &  \# Images &  \thead{\# Instances \\ (Expert)} &  \thead{\# Instances \\ (MTurk)} &  Median IoU &  \thead{\# Instances per Image \\ (Expert)} &  \thead{\# Instances per Image \\ (Mturk)} \\
\midrule
Zwarte Gouw                     &        80 &                  1,557 &                 1,050 &        0.62 &                              19 &                             13 \\
Overhoeks                       &       779 &                 15,270 &                11,296 &        0.68 &                              20 &                             15 \\
Haarlemmerbuurt Oost            &        99 &                  2,430 &                 1,847 &        0.74 &                              25 &                             19 \\
Zuidoostkwadrant Indische buurt &       702 &                 17,426 &                13,426 &        0.74 &                              25 &                             19 \\
G-buurt Noord                   &       104 &                  3,327 &                 2,523 &        0.74 &                              32 &                             24 \\
Johan Jongkindbuurt             &       108 &                  3,589 &                 2,652 &        0.75 &                              33 &                             25 \\
Leidsebuurt Noordwest           &        93 &                  1,880 &                 1,460 &        0.78 &                              20 &                             16 \\
Mercatorpark                    &       161 &                  5,746 &                 4,243 &        0.79 &                              36 &                             26 \\
\midrule
\textbf{All}                             &      \textbf{2,234} &                 \textbf{54,876} &                \textbf{41,238} &        \textbf{0.72} &                              \textbf{25} &                             \textbf{18} \\
\bottomrule
\end{tabular}
  }
\end{table*}

\begin{table*} 
\caption{Detection Performance on PanorAMS-clean, mAP@.5}
\label{tab:detection_per_topic_clean}
\centering
\begin{tabular}{llccccccccc}
\toprule
        & Trainset & Park &  Bus & Tram & Public Transport Stop & Playground & Advertising Column & Traffic Light & Public Toilet \\
\midrule
 Mask R-CNN & Clean (subset) & \textbf{3.5} &  4.5 &  4.9 &                   \textbf{3.8} &        \textbf{1.3} &                \textbf{5.7} &          11.3 &          \textbf{19.5}  \\
Cascade R-CNN & Clean (subset) &     3.2 &  \textbf{5.2} &  \textbf{5.5} &                   2.5 &        0.5 &                1.8 &          \textbf{13.8} &           0.0 \\
\bottomrule
\end{tabular}

\end{table*}

\section{Panoramic Distortion}

\begin{figure}
  \centering
  \includegraphics[width=0.48\textwidth]{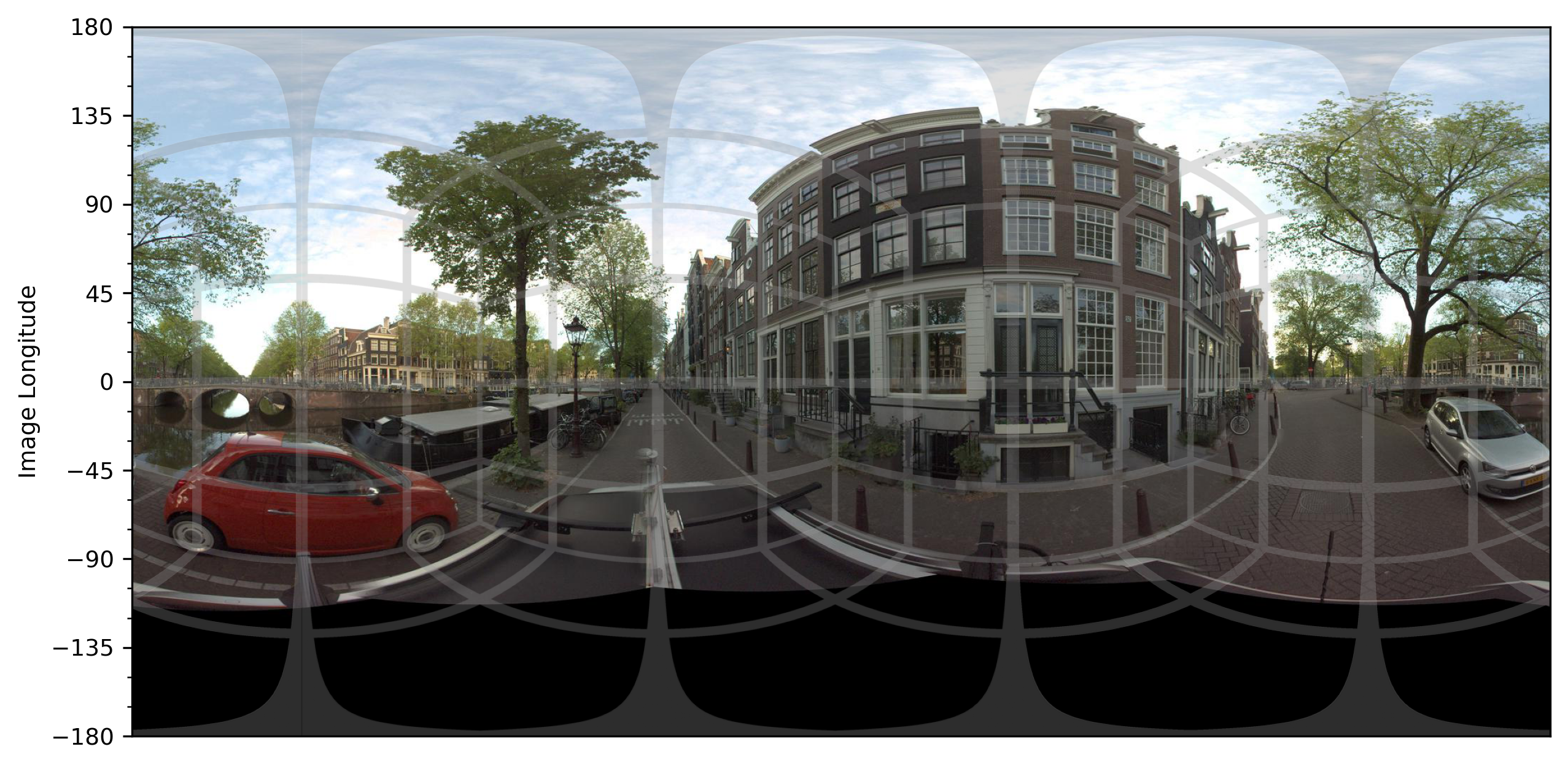}
  \caption{Street view image of Amsterdam with grid overlay revealing image distortion due to the panoramic projection.}
  \label{fig:panoramic_distortion_grid}
\end{figure}

\begin{figure}
  \centering
  \includegraphics[width=0.48\textwidth]{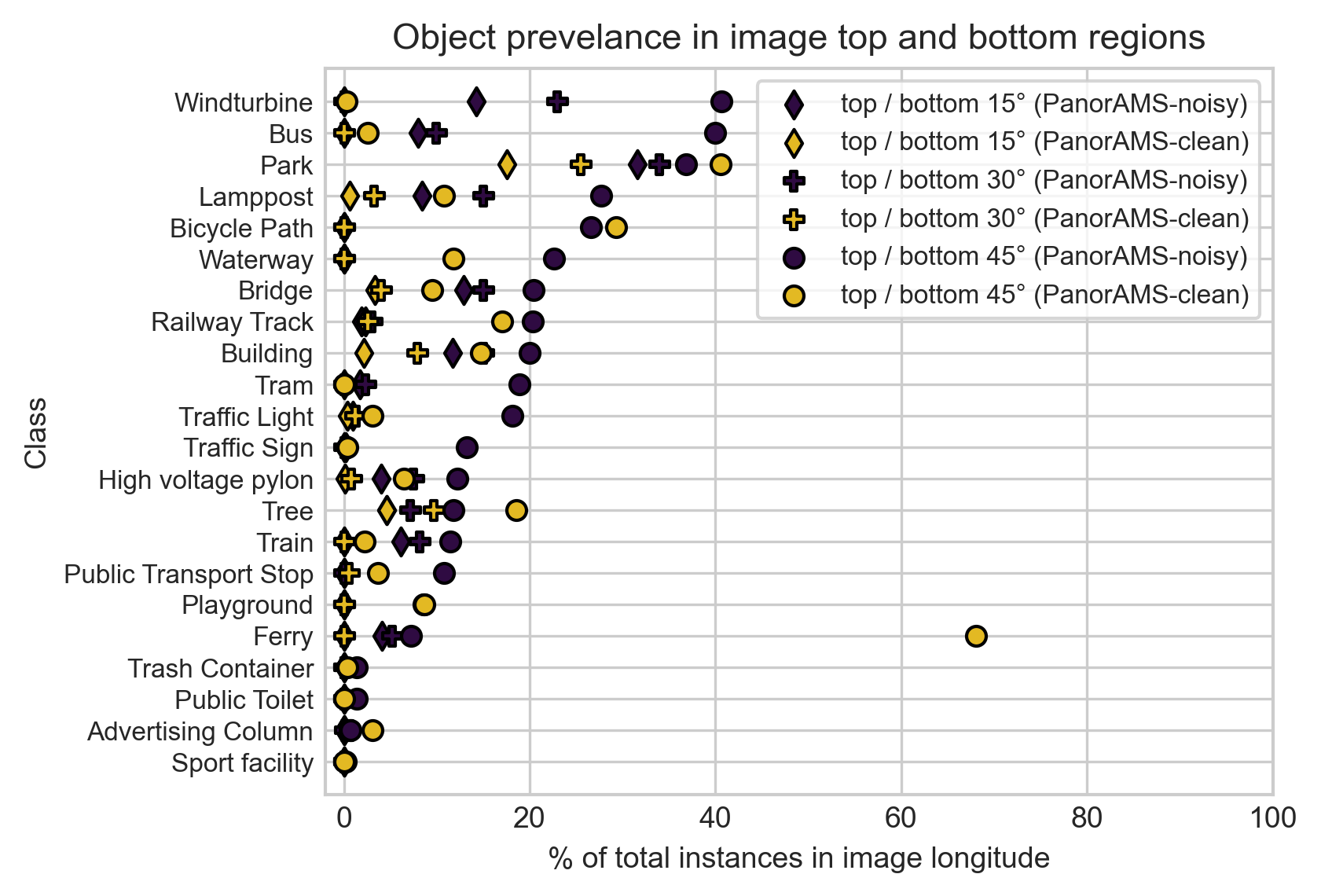}
  \caption{Percentage of instances per class that are present in the bottom and/or top regions of the image.}
  \label{fig:panoramic_distortion}
\end{figure}

Figure~\ref{fig:panoramic_distortion_grid} reveals the extent of distortion that occurs in our 360\textdegree{} panoramic street view images. At image longitude = 0 there is no image distortion, but as you move further along the y-axis towards image longitude = 180 the distortion increases. As indicated in Figure~\ref{fig:panoramic_distortion_grid}, there are generally no objects present in the bottom 60\textdegree{} of the image as this part of the image is usually covered by the moving vehicle taking the images. To provide further insight into how each object class is affected by panoramic distortion, we analyse the percentage of instances of a class that are present in the bottom and top regions of the image (cf. Figure~\ref{fig:panoramic_distortion}).

\section{Results}

\subsection{Classification}

\begin{figure}
  \centering
  \includegraphics[width=0.48\textwidth]{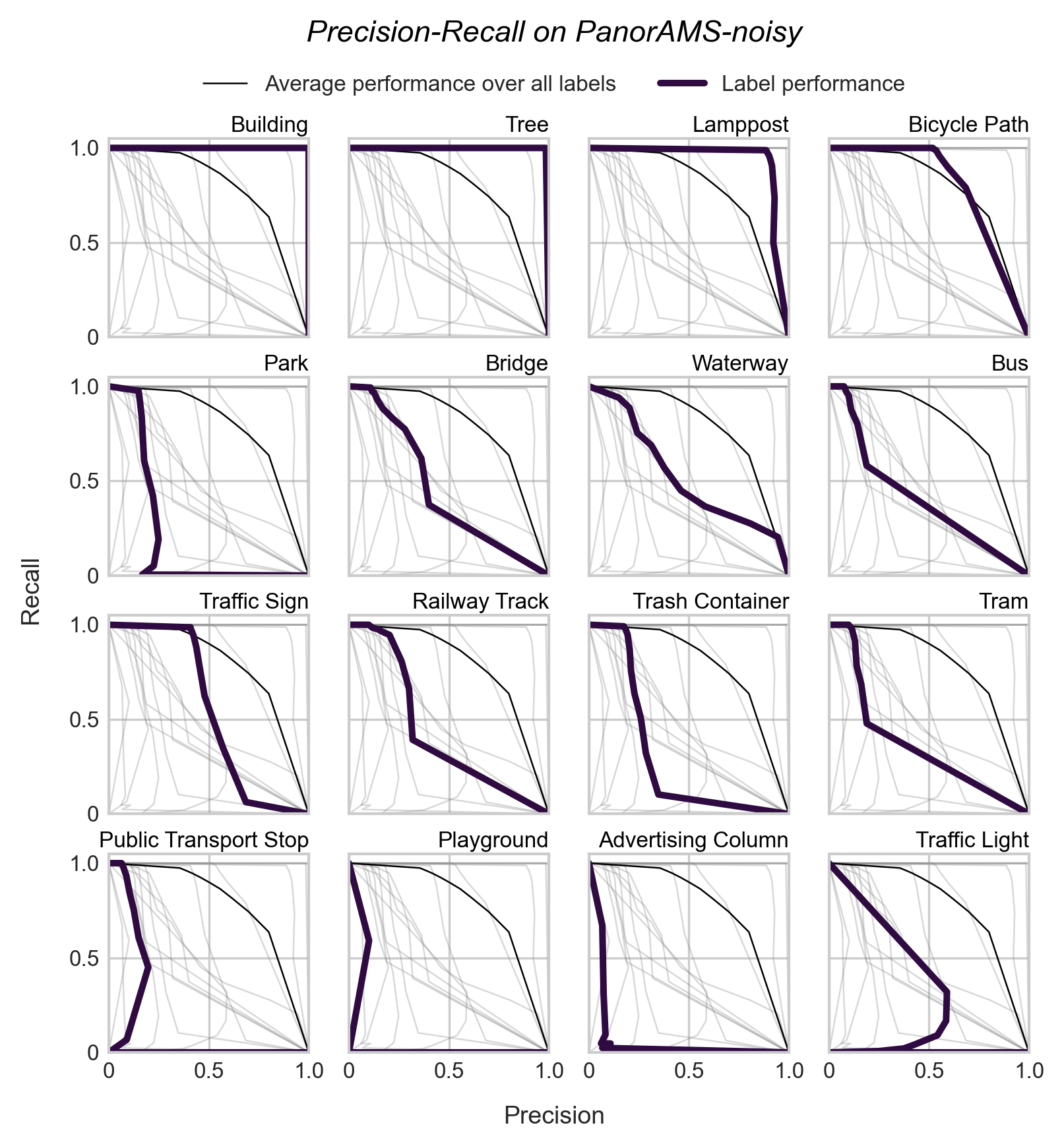}
  \caption{Precision-recall curve of classification performance of ResNet-50 trained on the Noisy (all) training set of PanorAMS-noisy.}
  \label{fig:pr_curve_noisy}
\end{figure}

\begin{figure}
  \centering
  \includegraphics[width=0.48\textwidth]{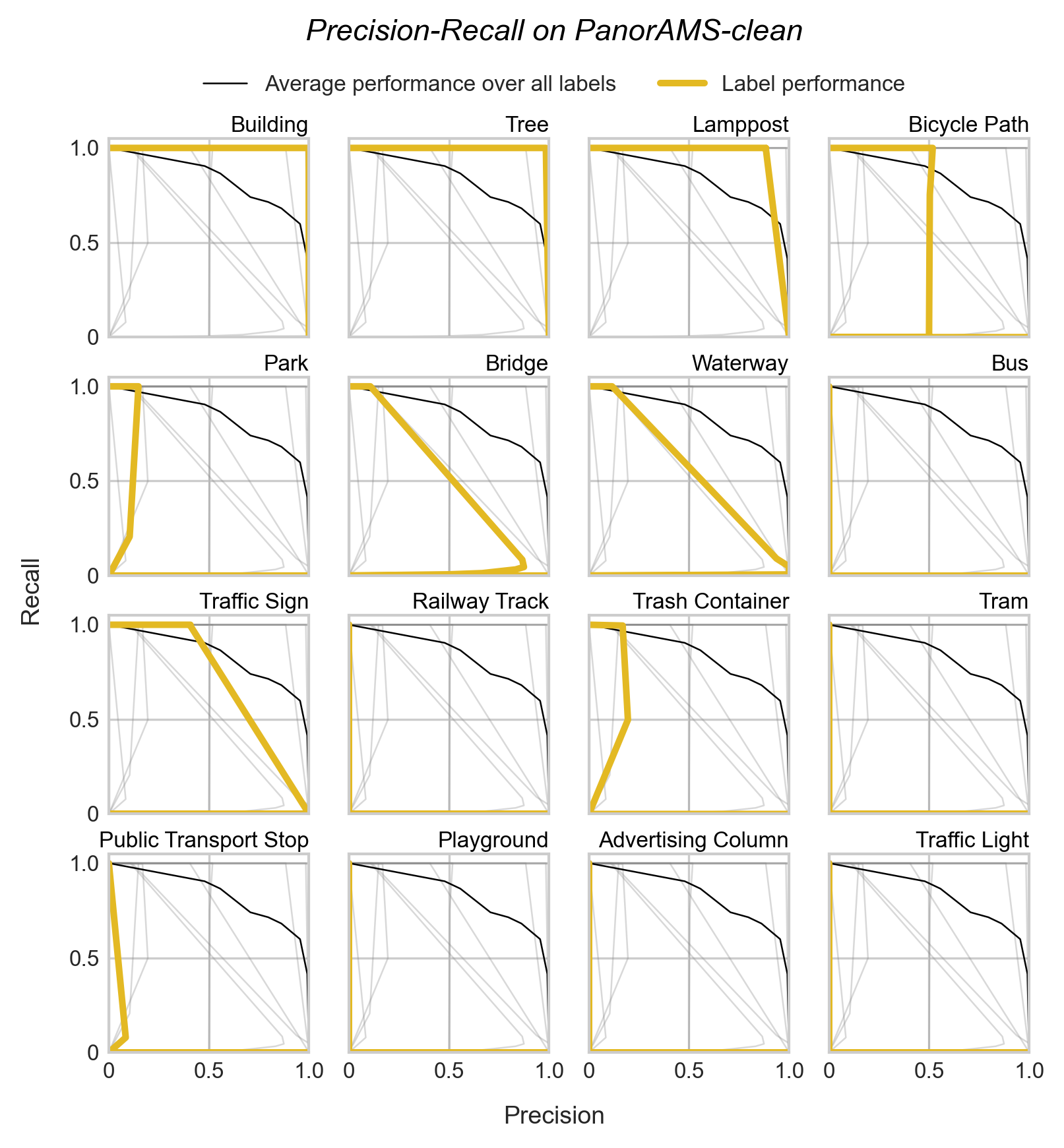}
  \caption{Precision-recall curve of classification performance of ResNet-50 trained on the Clean (subset) training set of PanorAMS-clean.}
  \label{fig:pr_curve_clean}
\end{figure}

In addition to the classification results presented and discussed in detail in Section VII-A of the main paper, we provide the precision-recall curves per class on, respectively, the PanorAMS-noisy  and PanorAMS-clean dataset (cf. Figures~\ref{fig:pr_curve_noisy} and ~\ref{fig:pr_curve_clean}). These results have been obtained by testing the networks on the Clean (test) split of PanorAMS-clean. 

\subsection{Detection}

In addition to the detection results presented and discussed in Section VII-B, Table~\ref{tab:detection_per_topic_clean} shows the detection results on PanorAMS-clean for difficult classes that the network could not learn using PanorAMS-noisy as the training set.